\documentclass[]{article}

\usepackage{cite}
\usepackage{amsmath,amssymb,amsfonts}
\usepackage{algorithmic}
\usepackage{graphicx}
\usepackage{textcomp}
\usepackage[table]{xcolor}
\usepackage{marvosym}
\usepackage{makecell}
\usepackage{multirow}
\usepackage[ruled,vlined,linesnumbered]{algorithm2e}
\usepackage[shortcuts]{extdash}
\usepackage[normalem]{ulem}
\usepackage{subcaption}
\usepackage{todonotes}
\usepackage{tabu}
\usepackage{booktabs}

\usepackage[margin=40mm]{geometry}
\providecommand{\keywords}[1]
{
  \small	
  \textbf{\textit{Keywords---}} #1
}

\begin{document}

\title{Prediction of the Facial Growth Direction\\is Challenging}
%
%
\author{
Stanisław Kaźmierczak$^1$,
Zofia Juszka$^2$, \\
Vaska Vandevska-Radunovic$^3$,
Thomas JJ Maal$^{4,5}$, \\
Piotr Fudalej$^{6,7,8}$,
Jacek Ma{\'n}dziuk$^1$ \\ \\

\small $^1$Faculty of Mathematics and Information Science, Warsaw University of Technology, \\
\small Warsaw, Poland, \texttt{\{s.kazmierczak,mandziuk\}@mini.pw.edu.pl} \\
\small $^2$Prof. Loster's Orthodontics, Krakow, Poland, \texttt{zofia.juszka@gmail.com} \\
\small $^3$Institute of Clinical Dentistry, University of Oslo, Oslo, Norway \\
\small $^4$Department of Oral and Maxillofacial Surgery 3D Lab, Radboud University Medical\\
\small Centre Nijmegen, Radboud Institute for Health Sciences, Nijmegen, The Netherlands \\
\small $^5$Department of Oral and Maxillofacial Surgery, Amsterdam UMC Location AMC  \\
\small and Academic Centre for Dentistry Amsterdam (ACTA), University of Amsterdam, \\
\small Amsterdam, The Netherlands \\
\small $^6$Department of Orthodontics, Jagiellonian University in Krakow, Krakow, Poland \\
\small \texttt{pfudalej@gmail.com} \\
\small $^7$Department of Orthodontics, Institute of Dentistry and Oral Sciences, \\
\small Palacký University Olomouc, Olomouc, Czech Republic \\
\small $^8$ Department of Orthodontics and Dentofacial Orthopedics, University of Bern, \\ 
\small Bern, Switzerland
}

\date{} 

\maketitle              

\begin{abstract}
Facial dysmorphology or malocclusion is frequently associated with abnormal growth of the face. The ability to predict facial growth (FG) direction would allow clinicians to prepare individualized therapy to increase the chance for successful treatment. Prediction of FG direction is a novel problem in the machine learning (ML) domain. In this paper, we perform feature selection and point the attribute that plays a central role in the abovementioned problem. Then we successfully apply data augmentation (DA) methods and improve the previously reported classification accuracy by 2.81\%. Finally, we present the results of two experienced clinicians that were asked to solve a similar task to ours and show how tough is solving this problem for human experts.


\end{abstract}

\section{Introduction}
\label{sec:Introduction}

The face plays numerous roles in humans – it is important in inter-human communication because it conveys emotions; it is used for recognition thanks to its unique three-dimensional morphology; moreover, it is considered a “locus of beauty” because it plays a key role in perception of physical attractiveness. Unfortunately, many pathological conditions affect appearance or function of the face. For example, orofacial cleft, the 2nd most prevalent congenital dysmorphology in humans, has a profound negative effect on facial development. As a result, only some patients have good-looking and/or well-functioning orofacial regions despite prolonged therapy. Also, atypical FG in many types of malocclusion hinders achievement of optimal results of treatment. Irrespective of the form of pathology or its severity, the type of FG, i.e., its direction, intensity, and duration, can be favorable, neutral, or unfavorable when viewed from the clinical perspective. Favorable FG takes place when elements of the face grow in a direction or with intensity promoting advantageous treatment outcome, while unfavorable growth occurs when growth characteristics does not facilitate treatment. A typical example is mandibular hypoplasia (i.e., when lower jaw is too small) – in some patients, a hypoplastic mandible grows significantly forward (favorable growth), whereas in others it grows considerably downward (unfavorable growth). Clearly, direction of growth of the mandible is associated with clinical success or failure. Therefore, it seems obvious that the ability to predict FG is of paramount importance for both patients and clinicians. It would allow to select more individualized therapy or more appropriate orthodontic appliance for an individual patient to increase the chance for successful treatment.

The main contribution of this work is fourfold:
\begin{itemize}
    \item addressing the problem of FG prediction with ML methods, which, to our knowledge, was analyzed only once before~\cite{kazmierczak2021Prediction} and improving the prediction score by 2.81\%; 
    \item showing that one feature -- the difference between the values at the age of 12 and 9 of the same measurement whose change between the 9th and 18th year of age is prognosticated -- plays a crucial role in the prediction process;
    \item successfully applying DA techniques to the analyzed problem;
    \item involving experienced clinicians to solve a similar problem and showing how difficult it is for human experts.
\end{itemize} 


\section{Related literature}
\label{sec:RelatedLiterature}

In our other article, we described in detail the attempts to date to predict craniofacial growth~\cite{kazmierczak2021Prediction}. In summary, none of the methods to date have been sufficiently effective when validated on a sample other than the one used to develop the method. 
ML methods have not been used for craniofacial growth prediction to date, despite the fact that recent years are replete with publications describing the use of ML in orthodontics and related areas~\cite{etemad2021machine,lo2021automatic,bianchi2020osteoarthritis}.

A recent scoping review ~\cite{mohammad2021machine} pointed out that artificial intelligence could assist in performing diagnostic assessments or determining treatment plans. However, the widespread clinical use of tools based on ML is far from complete. According to the authors, the most promising applications of these methods were landmark detection on lateral cephalograms, skeletal classification, and a tool designed to facilitate decision making in tooth extractions. 

In summary, application of ML methods in orthodontics and related areas has been growing fast in recent years. However, few challenging "orthodontic" problems have been solved with ML so far.

\section{Datasets}
\label{sec:Datasets}

In our previous study, we described the datasets in detail~\cite{kazmierczak2021Prediction}. In short, subjects from "Nittedal Growth Study" housed in the University of Oslo, Norway ~\cite{el1994longitudinal} and American Association of Orthodontists Foundation (AAOF) Craniofacial Growth Legacy Collection (Bolton-Brush, Burlington, Denver, Fels Longitudinal, Forsyth Twin, Iowa, Mathews, Michigan, and  Oregon collections) ~\cite{american1996growth} were included in this investigation. The respective growth studies recruited children born between the 1950s and mid-1970s who were recalled for periodical clinical examinations and radiographic (x-ray) imaging. In this study, lateral cephalograms (LC's), i.e., profile radiographs of the head, were used. The imaging technique and timing of radiographic registrations (i.e., at which age LC's were made) were attempted to be standardized only within the Growth Study but not between the Studies. Consequentially, LC's from different Growth Studies did not have the same magnification factor nor the timing of imaging. The vast majority of children had a West European ethnic background.

In our experiments, we intended to predict the change of the value SN/MP angle between the age of 9 and 18, on the basis of the cephalometric measurements at 9 and 12 years. Thus, we collected data from subjects who had LC's taken in timestamps as close as possible to 9, 12, and 18 years. The total number of samples included in this study was 639. Characteristics of particular age groups are depicted in Table~\ref{tab:datasetCharacteristic}. 

\begin{table}[b]
\centering
\caption{Characteristic of 9-, 12- and 18-year-olds.} \label{tab:datasetCharacteristic}
\begin{tabular}{m{2.2cm}m{2.2cm}m{1.5cm}m{1.6cm}}
\toprule
Group & Age & Minimum & Maximum \\
\midrule
\emph{9-year-olds} & $9.06 \pm 0.45$ & 6.00 & 10.92 \\
\emph{12-year-olds} & $12.07 \pm 0.39$ & 10.00 & 13.75\\
\emph{18-year-olds} & $17.41 \pm 1.71$ & 15.00 & 28.42\\
\bottomrule
\end{tabular}
\end{table}

\section{Experiments}
\label{sec:Experiments}

\begin{figure}[b!]
     \centering
     \begin{minipage}[t]{0.38\textwidth}
         \subfloat[] 
         {\includegraphics[width=\textwidth]{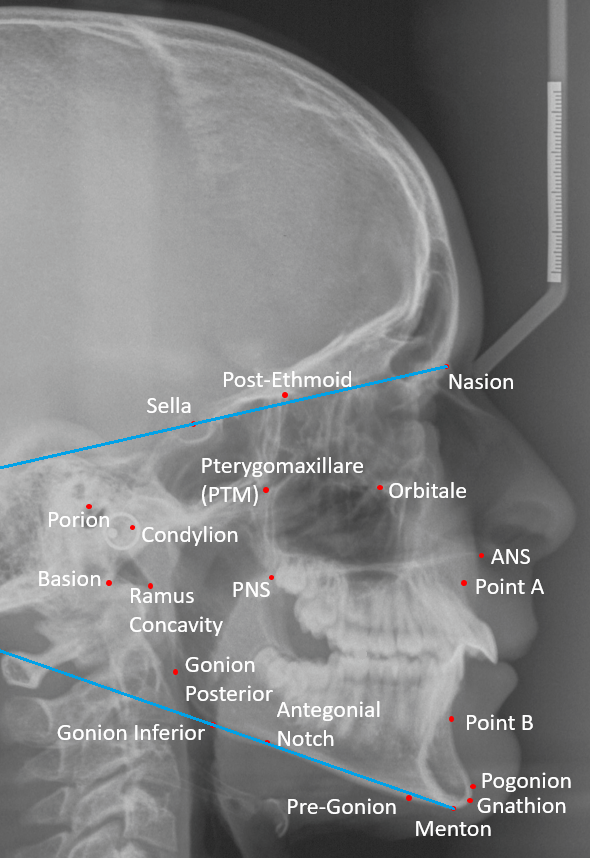}}
     \end{minipage}
     \hfill
     \begin{minipage}[b]{0.6\textwidth}
        \subfloat[]
        {\includegraphics[width=\textwidth]{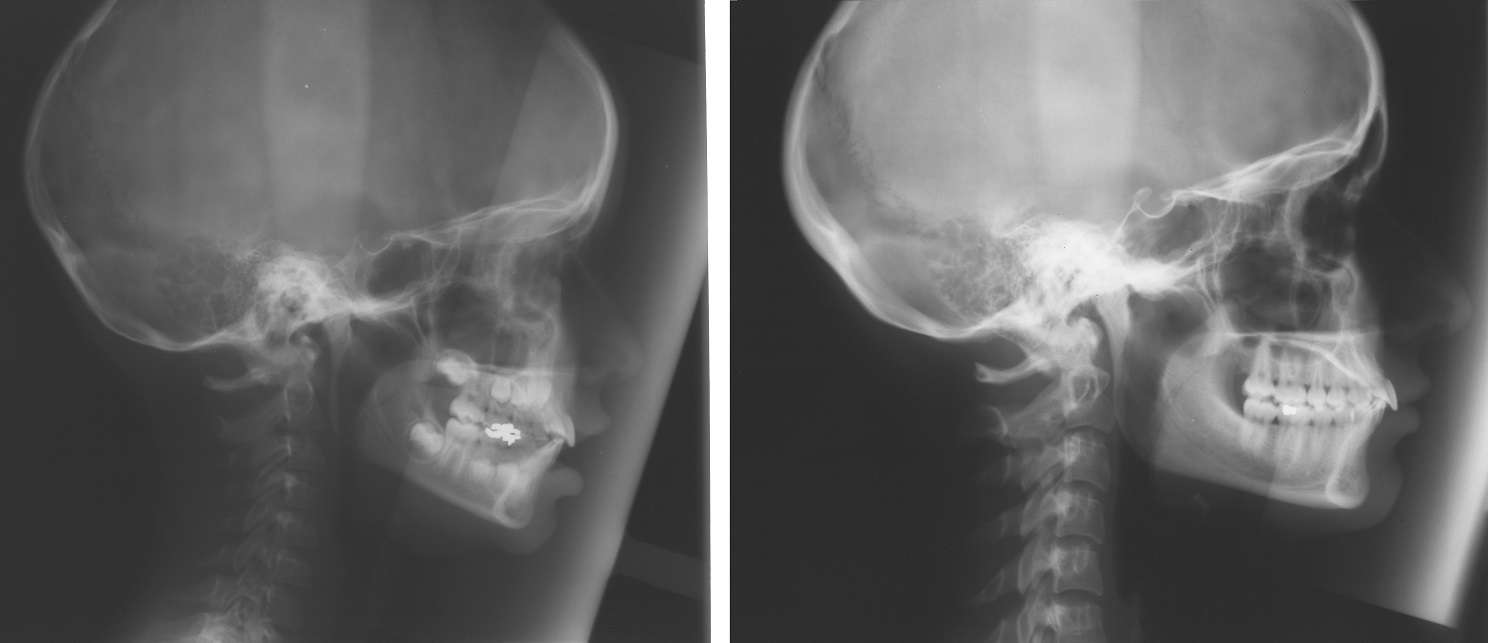}}\\
        \subfloat[]
        {\includegraphics[width=\textwidth]{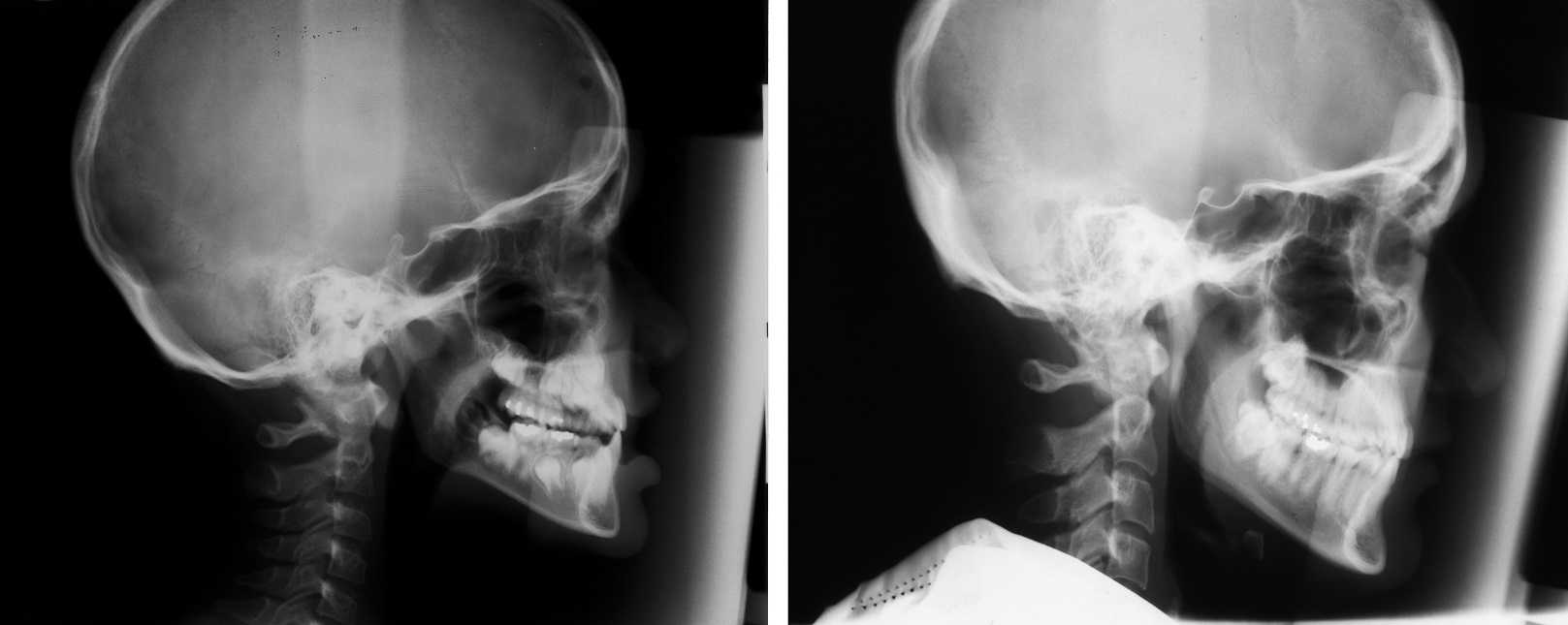}}
    \end{minipage}
    \caption{Sample cephalograms. Figure (a) depicts marked landmarks, their names, as well as the measurement used to  create  the  predicted variable -- SN/MP angle. Figures (b) and (c) present horizontal and vertical growth, respectively. They illustrate the faces of two people in the age of 9 (left) and 18 (right). }
    \label{fig:RTGExamples}
\end{figure}

To assess a subject, an orthodontic expert has to identify approximately 20 characteristic anatomic landmarks on LC. Some of them define angles which, from the clinical perspective, have special significance. As far as FG direction is concerned, there are no standardized measurements available in the literature to evaluate FG. In this paper, we focused on the prediction of the change of SN/MP, which is specified as the angle between Sella-Nasion and Menton-Gonion Inferior. The predicted variable is defined by subtracting the value of SN/MP at the age of 9 from SN/MP value at the age of 18. Categorization is conducted in the following way:
\begin{itemize}
\item first class, \emph{horizontal growth}, is composed of samples whose predicted value is lower than a standard deviation from the mean;
\item second, most frequent, class, \emph{mixed growth}, contains instances that are in the scope of one standard deviation from the mean;
\item third class, \emph{vertical growth}, constitutes samples with a predicted value greater than a standard deviation from the mean.
\end{itemize}
As a result, 68.23\% of instances belong to the most frequent class (MFC). Let us mark the categorized predicted variable as SN-MP(18-9).
Fig.~\ref{fig:RTGExamples}a illustrates an LC with landmarks and SN/MP angle.
Fig.~\ref{fig:RTGExamples}b and~\ref{fig:RTGExamples}c present subjects with horizontal and vertical growth, respectively.

\subsection{Task complexity}
Before moving to the experiment part, there are a couple of points worth noting that make solving the undertaken problem hard. The feature set is composed of either transformed landmarks or angles formed from these landmarks. Both approaches involve manual landmarking, which poses a source of noise since not every anatomical landmark location is unambiguous~\cite{perillo2000effect}. Additionally, the LC's included in this study were collected using different x-ray devices, which increases the non-uniformity of the data. Moreover, the image scale is diversified. Since it is impossible to determine the scaling factor, models may become confused when processing instances that do not reflect real growth properly. Further difficulties are related to the low number of dataset instances, which makes any deep architecture prone to overfitting and class imbalance. Last but not least, it is worth pointing out that we are restricted to work with 2D LC's due to the ethical issues described in~\cite{kazmierczak2021Prediction}.

\subsection{Models parametrization}
\label{sec:ModelsParametrization}
Let us introduce the following abbreviations for the sake of conciseness in the presentation:
\begin{itemize}
    \item MLP, MLP($n$), MLP($n_1, n_2$) stands for a perceptron with no hidden layer, one $n$-neuron hidden layer, and two hidden layers containing $n_1$ and $n_2$ neurons, respectively;
    \item NN($k$) is a nearest neighbors classifier that takes $k$ neighbors into account;
    \item XGB($r$) is an acronym XGBoost algorithm with $r$ boosting iterations;
    \item RF($t$) refers to a random forest containing $t$ trees;
    \item SVM pertains to Support Vector Machine classifier;
    \item LR corresponds to a logistic regression algorithm performing  $2\,000$ iterations at the maximum;
    \item DT relates to a decision tree. 
\end{itemize}

As far as neural networks are concerned, the input and each hidden layer applied ReLU, whereas the last layer used softmax. 20\% of training data were set aside and served as a validation set. The batch size was set to the number of training samples and the training could last no longer than 10\,000 epochs. After 50 epochs with no decrease of the loss function on the validation set, the training was ceased and parameters from the best epoch were brought back. Adam~\cite{kingma2014adam} served as the cross-entropy loss function optimizer. The remaining hyperparameters were left at their default settings, which were utilized in Keras 2.3.1 in terms of neural networks and scikit-learn 0.23.2 for all other models. Stratified 5-fold cross-validation was performed 20 times for each configuration presented in Section~\ref{sec:FeatureSelection} (generating 100 results) and 100 times for each setup provided in Section~\ref{sec:DataAugmentation} (yielding 500 results). In all experimental configurations, the following models were experimented: MLP, MLP(20), MLP(50), MLP(100), MLP(50, 10), MLP(50, 20), MLP(50, 50), SVM, LR, DT, NN(3), NN(5), RF(100), RF(300), XGB(100), XGB(300).

\subsection{Feature selection}
\label{sec:FeatureSelection}
In our experiments, the features are in tabular form (we comment on images in Section~\ref{sec:Conclusions}). We analyze the following three types of attributes: \emph{cephalometric} (ceph), \emph{Procrustes} (proc), and \emph{transformed} (trans). All of them are based on landmarks marked in LC's. Cephalometric features mainly consist of angles which are broadly used in the cephalometric analysis~\cite{leonardi2008automatic}. Images on which landmarking is based are in different scales. Moreover, the faces they depict are not centralized and are variously rotated. Thus, to create Procrustes features, we normalized the landmark coordinates so that they satisfy the following criteria for each face:
\begin{itemize}
\item arithmetic mean position of all the landmarks is placed at (0, 0);
\item total distance between all converted coordinates and (0, 0) equals one;
\item a sum of squared distances between a specific landmark and its mean position over all images is minimized across all landmarks and images.
\end{itemize}
The third set of features, transformed coordinates, is created by moving all raw landmarks so that the Sella point is positioned at (0, 0). Justification of such transformation and its visualization are presented in~\cite{kazmierczak2021Prediction}. The total number of cephalometric features from one timestamp amounts to 15, Procrustes -- 40, and transformed -- 40. Besides, the subject's age is always incorporated in the attribute set. In our experiments, we consider the features being the above-mentioned values from the 9th (9) and 12th (12) year of age, as well as the difference between the respective values at the age of 12 and 9 (12-9).

As it was shown in~\cite{kazmierczak2021Prediction}, there is a couple of sources of noise that affect our dataset. Thus, we decided to perform feature selection to check which attributes are meaningful from the prediction perspective. Since SN/MP(12-9) (difference between the values at the age of 12 and 9 of the same measurement whose change between the 9th and 18th year of age is prognosticated) is noticeably correlated with the predicted variable~\cite{kazmierczak2021Prediction}, we supposed that a relatively small set of attributes containing SN/MP(12-9) could give positive results. We started with the empty feature set and operated in a similar way to forward feature selection~\cite{blanchet2008forward}. 

First, we tested all types of models described in Section~\ref{sec:ModelsParametrization} built on any (but only one) attribute. Table~\ref{tab:topModels}a shows ten best configurations (model, feature, feature type), along with the obtained accuracy. All top models were built on SN/MP(12-9) attribute. MLP(100) outperformed the best model reported in~\cite{kazmierczak2021Prediction} by 1.12 percentage points. Additionally, the best one-feature model which is not built on SN/MP(12-9), achieved only 68.87\%. These two observations show, on the one hand how important SN/MP(12-9) attribute is, and on the other hand, how much noise is carried by many other features. 

Second, we tested all model types built on two features, out of which one is SN/MP(12-9). Table~\ref{tab:topModels}b depicts ten best configurations. Y Antegonial Notch(12\=/9) proved to be valuable. Obtained accuracy increased by 1.27 percentage points in comparison to the best one-feature model. However, only logistic regression was able to achieve such gain. It is also worth pointing out that nine out of ten models were built on the transformed coordinates and in the case of the top five models, the second feature was the difference between the corresponding values at the age of 12 and 9.

Finally, all models were examined on all possible three-feature sets, out of which two, SN/MP(12-9) and Y Antegonial Notch(12-9), were fixed and established by the best two-feature score. Table~\ref{tab:topModels}c shows the best configurations. Neither three-attribute model has obtained statistically significantly (\textit{t}-Test, \textit{p}-value equal to 0.05) higher accuracy than the best two-attribute model.
\begin{table}[htbp]
    \centering
    \caption{Top models built on one, two, and three features.} \label{tab:topModels}
        \begin{subtable}{0.875\linewidth}
            \caption{One-feature models}
            \begin{tabular}{m{0.7cm}m{2.2cm}m{5.3cm}m{2.0cm}}
            \toprule
            No. & Model & Feature and its type & Accuracy [\%] \\
            \midrule
            1 & MLP(100) & SN-MP(12-9), ceph & $72.37 \pm 1.87$\\
            2 & MLP(50) & SN-MP(12-9), ceph & $72.28 \pm 2.00$\\
            3 & MLP(50, 50) & SN-MP(12-9), ceph & $72.25 \pm 1.93$\\
            4 & MLP(20) & SN-MP(12-9), ceph & $72.24 \pm 2.02$\\
            5 & MLP(50, 20) & SN-MP(12-9), ceph & $72.17 \pm 1.93$\\
            6 & LR & SN-MP(12-9), ceph & $71.95 \pm 2.12$\\
            7 & SVM & SN-MP(12-9), ceph & $71.92 \pm 2.01$\\
            8 & MLP(50, 10) & SN-MP(12-9), ceph & $71.86 \pm 2.04$\\
            9 & MLP & SN-MP(12-9), ceph & $71.43 \pm 1.96$\\
            10 & XGB(100) & SN-MP(12-9), ceph & $71.22 \pm 2.61$\\
            \bottomrule
            \end{tabular}
        \end{subtable}%
        
        \vspace{0.2 cm}
        \begin{subtable}{0.875\linewidth}
            \caption{Two-feature models}
            \begin{tabular}{m{0.7cm}m{2.2cm}m{5.3cm}m{2.0cm}}
            \toprule
            No. & Model & $2^{nd}$ feature and its type & Accuracy [\%] \\
            \midrule
            1 & LR & Y Antegonial Notch(12-9), trans & $73.64 \pm 2.02$\\
            2 & LR & Y Gonion Inferior(12-9), trans & $72.64 \pm 2.05$\\
            3 & LR & X Gonion Inferior(12-9), trans & $72.51 \pm 2.14$\\
            4 & LR & Y Nasion(12-9), trans & $72.50 \pm 2.17$\\
            5 & MLP & Y Antegonial Notch(12-9), trans & $72.49 \pm 2.50$\\
            6 & LR & X Condylion(9), trans & $72.44 \pm 2.19$\\
            7 & MLP(100) & X Sella(9), trans & $72.43 \pm 1.88$\\
            8 & LR & SNA(9), ceph & $72.43 \pm 1.96$\\
            9 & LR & Y Gonion Posterior(12-9), trans & $72.43 \pm 2.08$\\
            10 & LR & X Point B(12-9), trans & $72.43 \pm 2.23$\\
            \bottomrule
            \end{tabular} 
        \end{subtable}%
        
        \vspace{0.2 cm}
        \begin{subtable}{0.875\linewidth}
            \caption{Three-feature models}
            \begin{tabular}{m{0.7cm}m{2.2cm}m{5.3cm}m{2.0cm}}
            \toprule
            No. & Model & $3^{rd}$ feature and its type & Accuracy [\%] \\
            \midrule
            1 & LR & SN/PP(12), ceph & $73.67 \pm 2.18$\\
            2 & LR & Y Post Ethmoid(12), proc & $73.65 \pm 2.02$\\
            3 & LR & X Nasion(12), proc & $72.65 \pm 2.02$\\
            4 & LR & Y Gonion Posterior(12-9), proc & $73.65 \pm 2.02$\\
            5 & LR & Y Gonion Inferior(12-9), proc & $73.65 \pm 2.03$\\
            6 & LR & Y Pre Gonion(12), proc & $73.65 \pm 2.04$\\
            7 & LR & AFH:PFH(12-9), ceph & $73.65 \pm 2.09$\\
            8 & LR & X Sella(9), trans & $73.64 \pm 2.02$\\
            9 & LR & Y Sella(9), trans & $73.64 \pm 2.02$\\
            10 & LR & X Nasion(9), trans & $73.64 \pm 2.02$\\
            \bottomrule
            \end{tabular}
        \end{subtable}%
\end{table}

\subsection{Data augmentation}
\label{sec:DataAugmentation}
Having found a strong configuration, logistic regression built on SN/MP(12-9) and Y Antegonial Notch(12-9) attributes, we attempted to improve the classification accuracy further by applying DA methods. Since the analyzed problem is imbalanced and features are in tabular form, we focused on the SMOTE~\cite{chawla2002smote} technique and its extensions. When augmenting instances of class~$C$, SMOTE randomly selects a sample $s$ of class $C$ and generates new instances in line segments connecting s and its $k$ nearest neighbors of class $C$ in the feature space.

Methods based on SMOTE but employing selective synthetic sample generation include Borderline-SMOTE, SVM-SMOTE, ADASYN, and K-means SMOTE. Borderline-SMOTE~\cite{han2005borderline} is based on the fact that instances located on the decision boundary or next to it are more likely to be misclassified than the ones lying far from the borderline. Thus, the algorithm generates a new sample only if it lies on the decision boundary. SVM-SMOTE~\cite{nguyen2011borderline} is very similar to Borderline-SMOTE, but SVM is applied to form a borderline instead of k-nearest neighbors. ADASYN~\cite{he2008adasyn} is also based on SMOTE and generates synthetic data. However, it concentrates on producing new instances around those located in low-density areas. K-means SMOTE~\cite{douzas2018improving}, in turn, creates new samples based on the density of each cluster found by the k-means algorithm.

\begin{figure}[]
    \captionsetup[subfigure]{justification=centering}
     \centering
     \begin{subfigure}[]{0.45\textwidth}
         \centering
         \includegraphics[width=\textwidth]{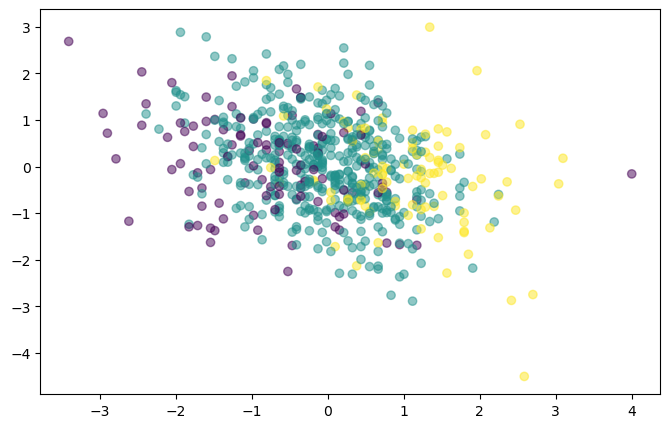}
         \caption{Original}
         \label{subfig:Original}
     \end{subfigure}
     \hspace{0.9 cm}
     \begin{subfigure}[]{0.45\textwidth}
         \centering
         \includegraphics[width=\textwidth]{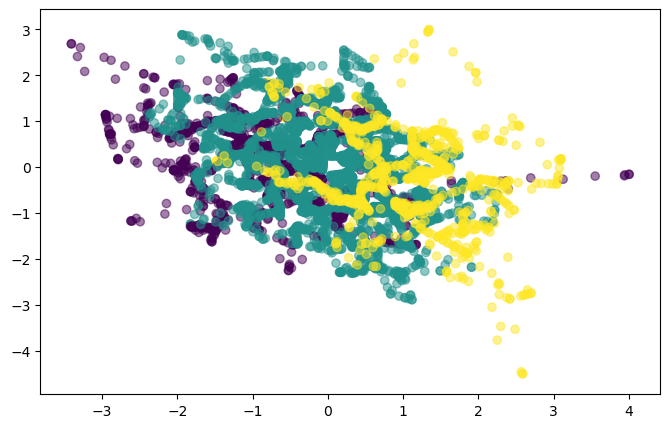}
         \caption{SMOTE}
         \label{subfig:SMOTE}
     \end{subfigure}
     
     \vspace{0.2 cm}
     \begin{subfigure}[]{0.45\textwidth}
         \centering
         \includegraphics[width=\textwidth]{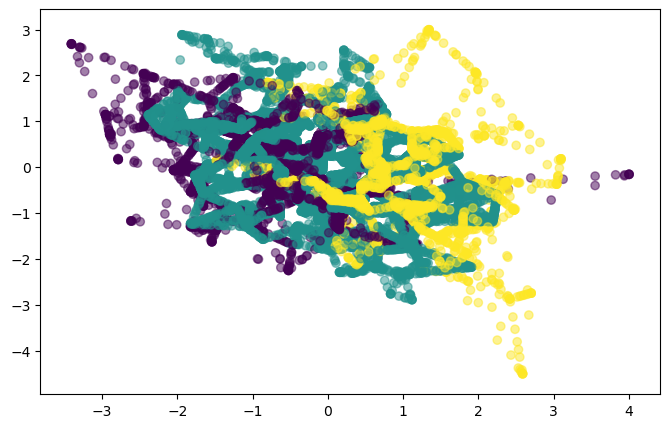}
         \caption{Borderline-SMOTE}
         \label{subfig:Borderline}
     \end{subfigure}
     \hspace{0.9 cm}
     \begin{subfigure}[]{0.45\textwidth}
         \centering
         \includegraphics[width=\textwidth]{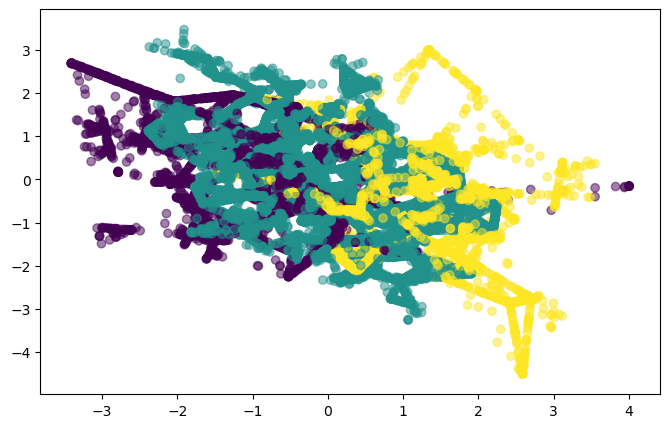}
         \caption{SVM-SMOTE}
         \label{subfig:SVM}
     \end{subfigure}

     \vspace{0.2 cm}
     \begin{subfigure}[]{0.45\textwidth}
         \centering
         \includegraphics[width=\textwidth]{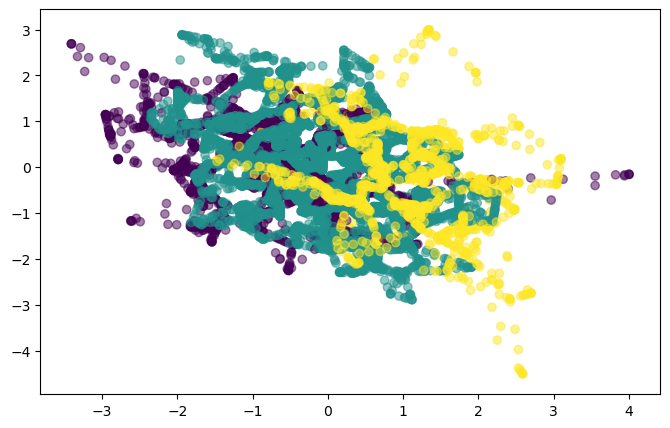}
         \caption{ADASYN}
         \label{subfig:ADASYN}
     \end{subfigure}
     \hspace{0.9 cm}
     \begin{subfigure}[]{0.45\textwidth}
         \centering
         \includegraphics[width=\textwidth]{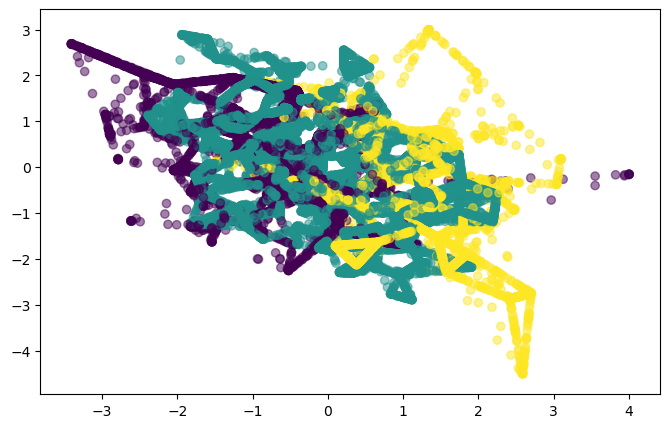}
         \caption{K-means SMOTE}
         \label{subfig:KMEANS}
     \end{subfigure}

     \vspace{0.2 cm}
     \begin{subfigure}[]{0.45\textwidth}
         \centering
         \includegraphics[width=\textwidth]{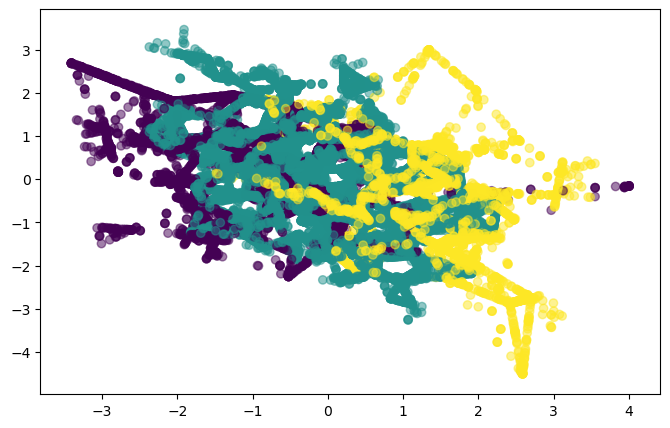}
         \caption{SMOTE-Tomek}
         \label{subfig:Tomek}
     \end{subfigure}
     \hspace{0.9 cm}
     \begin{subfigure}[]{0.45\textwidth}
         \centering
         \includegraphics[width=\textwidth]{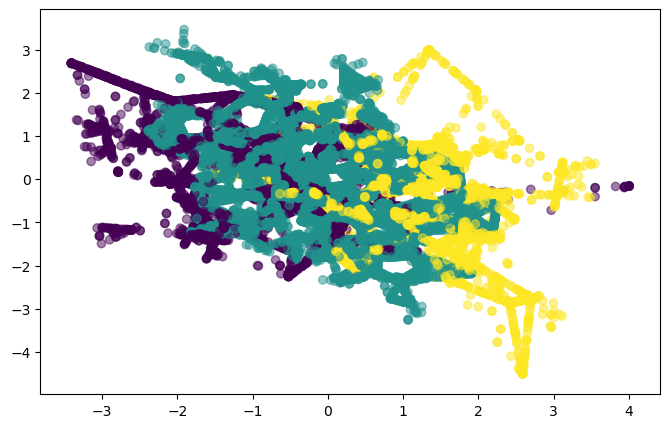}
         \caption{SMOTEENN}
         \label{subfig:ENN}
     \end{subfigure}
    \caption{SMOTE and its extentions; \textit{x}-axis -- standarized SN/MP(12-9), \textit{y}-axis -- standarized Y Antegonial Notch(12-9), augmentation factor -- 10.}
    \label{fig:visualization}
\end{figure}

SMOTE can create noisy instances when interpolating new points between marginal outliers and inliers. To deal with that issue, we employ two methods that clean a space resulting from DA. Both techniques start with performing SMOTE, but they differ in the second phase. The first method, SMOTE-Tomek Links~\cite{batista2004study}, involves removing instances of the majority class which are closest to samples of the minority class. The second, SMOTE-ENN~\cite{batista2003balancing}, applies the edited nearest-neighbors algorithm. After running SMOTE, it erases instances that do not match \emph{enough} with their neighborhood. Fig.~\ref{fig:visualization} visualizes the aforementioned methods applied to the analyzed dataset.

At the beginning of the experiments related to DA, we tried to enlarge the underrepresented class by a greater factor than the majority class. However, it turns out that when disturbing the original class proportion, the classification accuracy suffers. Thus, in further analysis, we stick to scenarios in which all three classes are augmented by the same factor. 

\begin{figure}[t!]
\centerline{\includegraphics[width=1.0\textwidth]{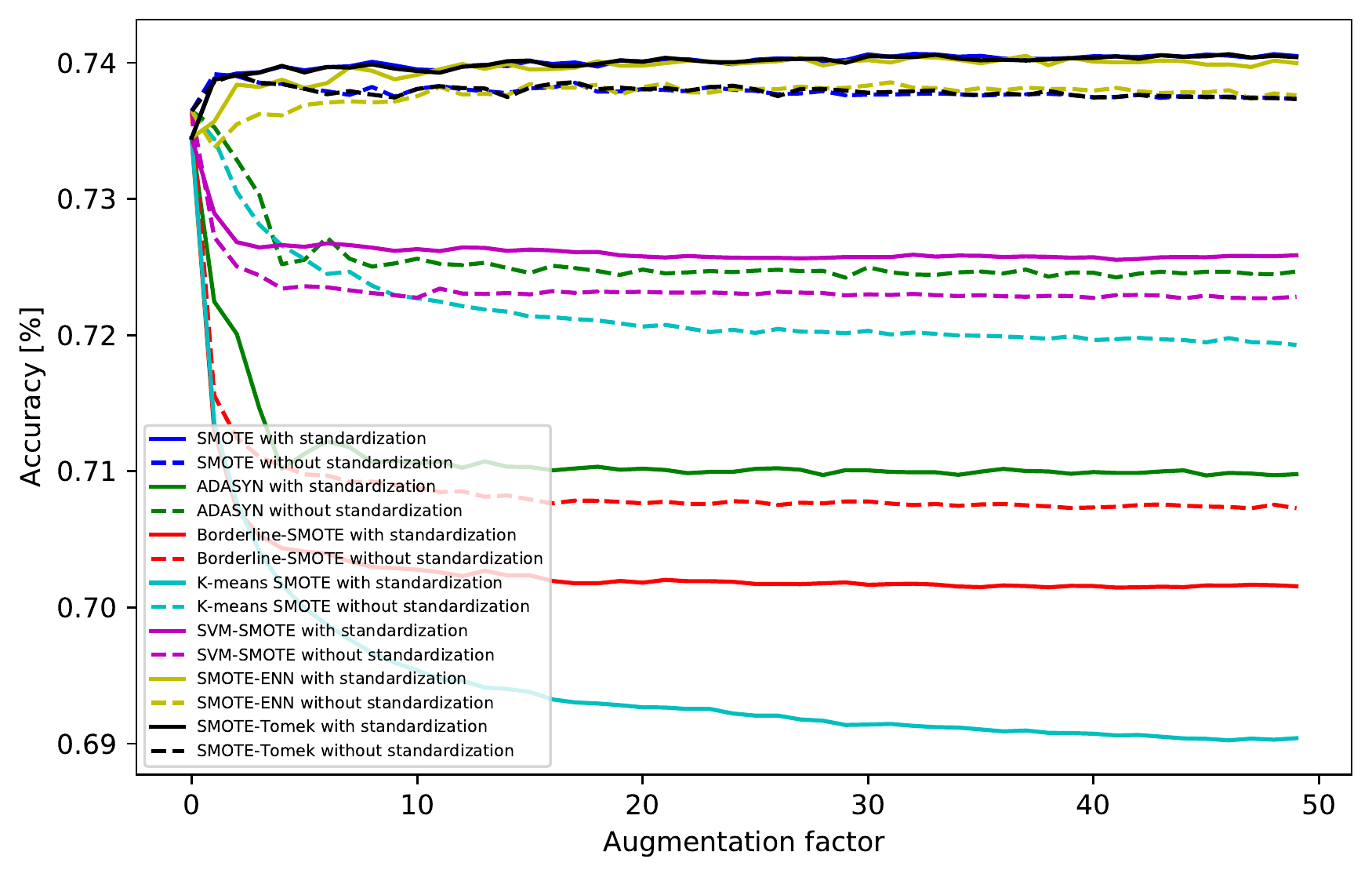}}
\caption{Classification accuracies obtained by different SMOTE-related DA methods.}
\label{fig:SMOTE}
\end{figure}

All aforementioned DA methods are run on the original SN/MP(12-9) and Y Antegonial Notch(12-9) features, as well as their standardized values. In all experiments, we utilize logistic regression, which obtained the best results on the above attributes. Fig.~\ref{fig:SMOTE} shows classification accuracy as a function of the augmentation factor, which tells how many times the generated dataset is larger than the original one. The training dataset is composed of two parts -- original and augmented instances.

\begin{figure}[htbp!]
\centerline{\includegraphics[width=1.0\textwidth]{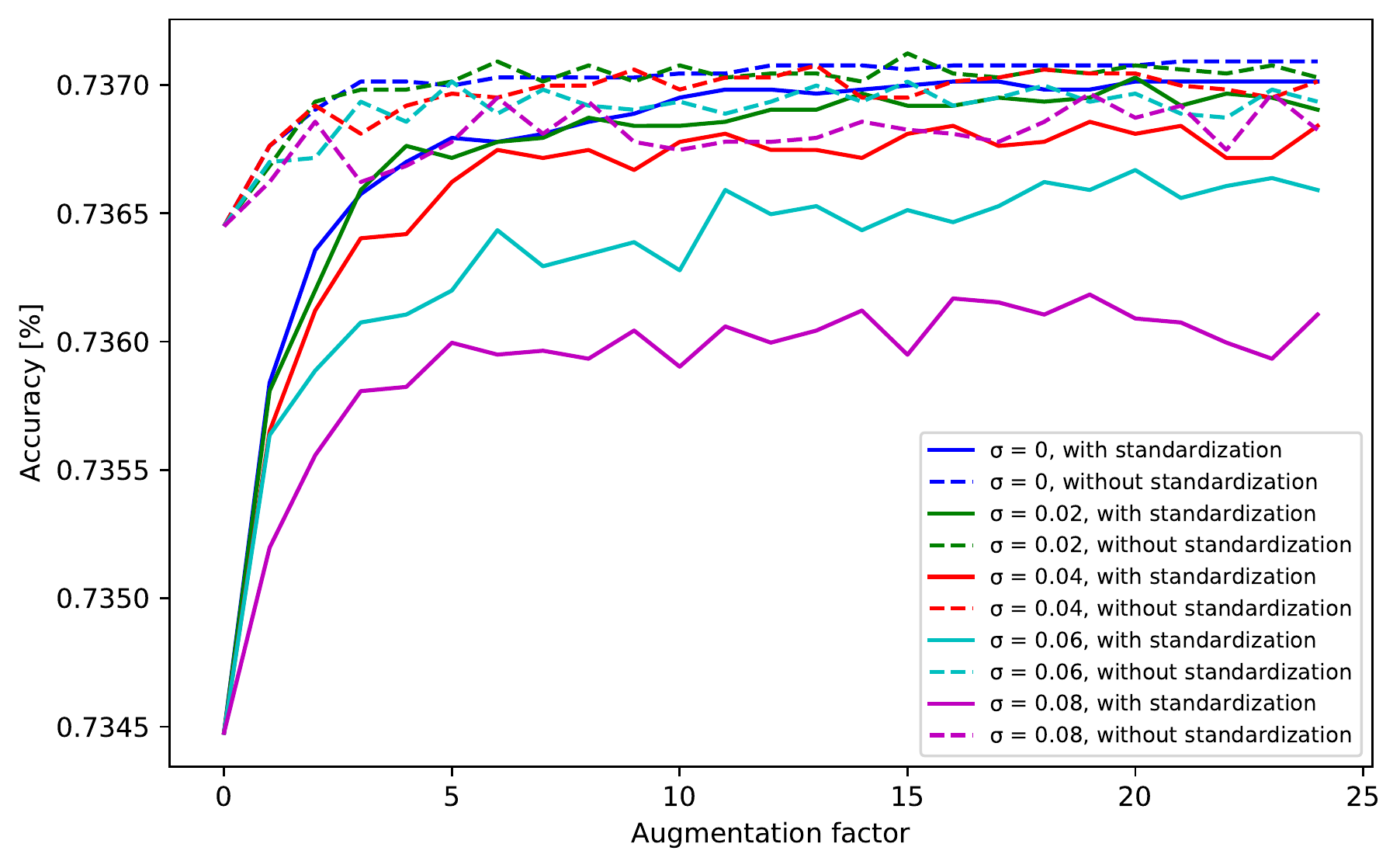}}
\caption{Results achieved by addition of Gaussian noise.}
\label{fig:Gauss}
\end{figure}

The baseline for further analysis constitutes accuracies obtained on non-augmented datasets -- 73.64\% for original features and 73.45\% on standardized. What can be observed in Fig.~\ref{fig:SMOTE}, only SMOTE and SMOTE with additional undersampling, SMOTE-Tomek and SMOTE-ENN, managed to outperform baseline. Moreover, the results achieved on standardized data are higher than those on raw data. Models built on standardized data reach a plateau slightly above 74.00\% when the augmentation factor is about 20 and achieve the highest accuracy, 74.06\%, for several larger values of the augmentation factor. Conversely, models constructed on the original data start to decrease very slightly then. Moreover, for most values of augmentation factor, SMOTE, SMOTE-TOMEK, and SMOTE-ENN applied to standardized data, achieve statistically significantly higher accuracy than the baseline (it is impossible to give the specific minimum value of accuracy to be viewed as statistically significantly different since the standard deviation varies a bit for different values of augmentation factor, but it oscillates about 73.90\%). We regard two mean accuracies to be statistically significantly different if the \textit{p}-value returned by \textit{t}-Test is less than 0.05. Finally, we tested DA based on Gaussian noise injection. Fig.~\ref{fig:Gauss} presents results for a different level of noise. Neither model obtained a result that could be regarded as statistically significantly higher than the baseline.

\subsection{Results of experts} 
To our knowledge,~\cite{kazmierczak2021Prediction} and this paper are the first studies investigating the prediction of FG with ML methods. Thus, we are unable to compare our outcome with any previous results. To check the level of difficulty of the problem and assess human-level performance, we asked two orthodontists with more than 10\=/year experience in cephalometric analysis to perform prediction. They were shown LC's taken at 9 years of age along with the results of cephalometric analysis as an input. Their task was to predict the growth category (categorized SN/MP(18-9) variable) for each subject. Experts obtained 40.33\% and 40.88\% of classification accuracy, respectively. Their prediction consistency (percentage of instances predicted as the same class) was 46.96\%. Results revealed how difficult it is for humans to predict the growth direction having 9-year-olds' data as an input.

\section{Conclusions and future work}
\label{sec:Conclusions}
This paper addresses FG prediction, which is a novel problem in the ML domain. The first interesting finding is that there is one feature, the difference between the values at the age of 12 and 9 of the same measurement whose change between the 9th and 18th year of age is prognosticated, that plays a crucial role in the prediction process. It is so important that models built solely on it outperform all other models tested in our previous work~\cite{kazmierczak2021Prediction}. The feature selection process reveals that there are many attributes redundant from the prediction perspective, and just two features constitute a suboptimal or even optimal attribute set. The broad range of tabular DA methods was further applied, which allowed us to gain another 0.4\% in classification accuracy. In all, the achieved score surpasses the previously reported classification accuracy by 2.81\%.  We also asked two experienced clinicians to perform predictions based on the data of 9-year-olds. Results unveiled how hard the problem is.

In terms of further research, one could pose a question about using convolutional neural networks, which achieve state-of-the-art results in many computer vision tasks and their direct application to our radiographic images (LC's). It is definitely worth checking, however, due to the high level of noise in LC's, the nuanced character of the problem, and the low volume of data, we do not expect spectacular results in this approach. Instead, we think of the application of some specific ensembling that will not combine models predicting the same variable but predictions of different measurements related to FG direction. Finally, we plan to look for some new data, not necessarily 2D LC's, that may render vital and carry predictive power.

\paragraph{Acknowledgments} Studies were funded by BIOTECHMED-1 project granted by Warsaw University of Technology under the program Excellence Initiative: Research University (ID-UB). We would like to thank the custodian of AAOF Craniofacial Growth Legacy Collection for the possibility to use the lateral cephalograms from Craniofacial Growth Legacy Collection.

\bibliographystyle{splncs04}
\bibliography{references}

\begin{thebibliography}{10}
\providecommand{\url}[1]{\texttt{#1}}
\providecommand{\urlprefix}{URL }
\providecommand{\doi}[1]{https://doi.org/#1}

\bibitem{american1996growth}
{American Growth Studies} (1996), \url{https://aaoflegacycollection.org/},
  online; accessed 02-Jun-2021

\bibitem{batista2003balancing}
Batista, G.E., Bazzan, A.L., Monard, M.C., et~al.: {Balancing Training Data for
  Automated Annotation of Keywords: a Case Study.} In: WOB. pp. 10--18 (2003)

\bibitem{batista2004study}
Batista, G.E., Prati, R.C., Monard, M.C.: {A Study of the Behavior of Several
  Methods for Balancing Machine Learning Training Data}. ACM SIGKDD
  Explorations Newsletter  \textbf{6}(1),  20--29 (2004)

\bibitem{bianchi2020osteoarthritis}
Bianchi, J., de~Oliveira~Ruellas, A.C., Gon{\c{c}}alves, J.R., Paniagua, B.,
  Prieto, J.C., Styner, M., Li, T., Zhu, H., Sugai, J., Giannobile, W., et~al.:
  Osteoarthritis of the temporomandibular joint can be diagnosed earlier using
  biomarkers and machine learning. Scientific Reports  \textbf{10}(1),  1--14
  (2020)

\bibitem{blanchet2008forward}
Blanchet, F.G., Legendre, P., Borcard, D.: Forward selection of explanatory
  variables. Ecology  \textbf{89}(9),  2623--2632 (2008)

\bibitem{chawla2002smote}
Chawla, N.V., Bowyer, K.W., Hall, L.O., Kegelmeyer, W.P.: {SMOTE: Synthetic
  Minority Over-sampling Technique}. Journal of Artificial Intelligence
  Research  \textbf{16},  321--357 (2002)

\bibitem{douzas2018improving}
Douzas, G., Bacao, F., Last, F.: Improving imbalanced learning through a
  heuristic oversampling method based on k-means and smote. Information
  Sciences  \textbf{465},  1--20 (2018)

\bibitem{el1994longitudinal}
El-Batouti, A., {\O}gaard, B., Bishara, S.E.: Longitudinal cephalometric
  standards for norwegians between the ages of 6 and 18 years. European Journal
  of Orthodontics  \textbf{16}(6),  501--509 (1994)

\bibitem{etemad2021machine}
Etemad, L., Wu, T.H., Heiner, P., Liu, J., Lee, S., Chao, W.L., Zaytoun, M.L.,
  Guez, C., Lin, F.C., Jackson, C.B., Ko, C.C.: Machine learning from clinical
  datasets of a contemporary decision for orthodontic tooth extraction.
  {Orthodontics \& Craniofacial Research}  (2021)

\bibitem{han2005borderline}
Han, H., Wang, W.Y., Mao, B.H.: {Borderline-SMOTE: A New Over-Sampling Method
  in Imbalanced Data Sets Learning}. In: International Conference on
  Intelligent Computing. pp. 878--887. Springer (2005)

\bibitem{he2008adasyn}
He, H., Bai, Y., Garcia, E.A., Li, S.: {ADASYN: Adaptive Synthetic Sampling
  Approach for Imbalanced Learning}. In: 2008 IEEE International Joint
  Conference on Neural Networks. pp. 1322--1328. IEEE (2008)

\bibitem{kazmierczak2021Prediction}
Ka{\'z}mierczak, S., Juszka, Z., Fudalej, P., Ma{\'n}dziuk, J.: Prediction of
  the facial growth direction with machine learning methods. arXiv preprint
  arXiv:2106.10464  (2021)

\bibitem{kingma2014adam}
Kingma, D.P., Ba, J.: {Adam: A Method for Stochastic Optimization}. arXiv
  preprint arXiv:1412.6980  (2014)

\bibitem{leonardi2008automatic}
Leonardi, R., Giordano, D., Maiorana, F., Spampinato, C.: Automatic
  cephalometric analysis: A systematic review. The Angle Orthodontist
  \textbf{78}(1),  145--151 (2008)

\bibitem{lo2021automatic}
Lo, L.J., Yang, C.T., Ho, C.T., Liao, C.H., Lin, H.H.: Automatic assessment of
  3-dimensional facial soft tissue symmetry before and after orthognathic
  surgery using a machine learning model: A preliminary experience. {Annals of
  Plastic Surgery}  \textbf{86}(3S),  S224--S228 (2021)

\bibitem{mohammad2021machine}
Mohammad-Rahimi, H., Nadimi, M., Rohban, M.H., Shamsoddin, E., Lee, V.Y.,
  Motamedian, S.R.: Machine learning and orthodontics, current trends and the
  future opportunities: A scoping review. American Journal of Orthodontics and
  Dentofacial Orthopedics  (2021)

\bibitem{nguyen2011borderline}
Nguyen, H.M., Cooper, E.W., Kamei, K.: {Borderline Over-sampling for Imbalanced
  Data Classification}. International Journal of Knowledge Engineering and Soft
  Data Paradigms  \textbf{3}(1),  4--21 (2011)

\bibitem{perillo2000effect}
Perillo, M., Beideman, R., Shofer, F., Jacobsson-Hunt, U., Higgins-Barber, K.,
  Laster, L., Ghafari, J.: Effect of landmark identification on cephalometric
  measurements: guidelines for cephalometric analyses. Clinical Orthodontics
  and Research  \textbf{3}(1),  29--36 (2000)

\end{thebibliography}

\end{document}